\documentclass[a4paper]{article}

\usepackage{amsfonts}
\usepackage{amsmath}

\usepackage{mathrsfs}
\usepackage{pst-all}
\usepackage{graphicx}
\usepackage{pslatex}

\psset{unit=0.4cm}

\def\DF#1{\textbf{#1}}

\def\FORALL{\bigwedge}
\def\EXISTS{\bigvee}

\def\Z{\mathbb{Z}}
\def\N{\mathbb{N}}
\def\R{\mathbb{R}}

\def\P{\mathcal{P}}

\def\T{\mathcal{T}}

\def\SUBSET{\subset}

\def\TO{\to}

\def\IFF{\Leftrightarrow}
\def\THEN{\Rightarrow}
\def\IF{\Rightarrow}
\def\ONLYIF{\Leftarrow}

\def\POT{\mathfrak{P}}

\def\TOP{\mathscr{T}}
\def\TOPP{\mathscr{U}}

\def\A{\mbox{\textnormal{A}}}
\def\U{\mbox{\textnormal{U}}}
\def\C{\mbox{\textnormal{C}}}
\def\dim{\mbox{\textnormal{dim}}}

\def\FORALL{\bigwedge}
\def\EXISTS{\bigvee}

\newtheorem{Satz}{Theorem}%[section]
\newtheorem{Def}{Definition}%[section]
\newtheorem{Lemma}{Lemma}%[section]
\newtheorem{Folg}{Corollary}%[section]

\newenvironment{Beweis}
{\paragraph{\textit{Proof.}}}
{\hspace{\stretch{1}}$\Box$\vspace{5mm}}
\def\qed{\hspace{\stretch{1}}$\Box$}

\def\SET#1#2{\{ {#1},\ldots,{#2}\} }
\def\ASET#1{\left\{ {#1} \right\} }
\def\COMP#1{{#1}^C}
\def\SETMINUS{\setminus}

\title{Good Pairs of Adjacency Relations in Arbitrary Dimensions}

\author{Martin H\"unniger}
\date{ }

\begin{document}
\maketitle

\pagestyle{plain}
\pagenumbering{roman}

\begin{abstract}
  In this text we show that the notion of a ``good pair'' that was introduced in the 
  paper \cite{paper01} has actually known models. We will show, how to choose cubical 
  adjacencies, the generalizations of the well known 4- and 8-neighborhood to
  arbitrary dimensions, in order to find good pairs. Furthermore, we give another proof for the
  well known fact that the Khalimsky-topology \cite{khalimsky} implies good pairs.
  The outcome is consistent with the known theory as presented by T.Y.~Kong, A.~Rosenfeld
  \cite{kongrose}, G.T.~Herman \cite{herman} and M.~Khachan et.al \cite{khachan}
  and gives new insights in higher dimensions.
\end{abstract}

\tableofcontents

\pagenumbering{arabic}

\section{Introduction}

In the text \cite{paper01} the author has given a new framework to define 
$(n-1)$-manifolds in $\Z^n$ together with a notion of ``good pairs'' of adjacency relations.
Such a good pair makes it possible for a $(n-1)$-manifold to satisfy a discrete
analog of the Theorem of Jordan-Brouwer. This Theorem is a generalization of the 
Jordan-curve Theorem, which states that every simple closed curve in $\R^2$ separates its 
complement in exactly two connected components and is itself the boundary of both of them.
Brouwer showed that the statement is true for simple $(n-1)$-manifolds in $\R^n$ for all
$n\ge2$. It has been an open question since the beginnings of digital image analysis, if this 
is true in a discrete setting, so to speak in $\Z^n$. 

\begin{figure}[htb]
  \begin{center}
    \includegraphics{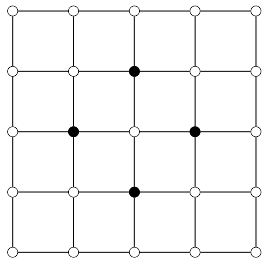}
  \end{center}
\caption[Initial problem]{\small Depending on the adjacency relations we use for the 
black and the white points, respectively, the set of black points is connected 
(8-adjacency) or disconnected (4-adjacency). Also the set of white points may be connected
(8-adjacency) or disconnected (4-adjacency). Only 4-adjacency is depicted.}
\label{Pic:ausgangs_problem}
\end{figure}

As the figure \ref{Pic:ausgangs_problem} shows, it is not even clear what a simple 
closed curve should look like in a discrete setting. And really, this depends on 
the adjacency we impose
on the points of $\Z^n$. We also see from the figure, that it is not enough to use only
one adjacency for the base-set (background / white points) and the objects 
(foreground / black points), we have to use pairs of them. Unfortunately, not every pair of
adjacencies is suitable because some even fail to make a ($n-1$)-manifold out of the neighbors
of a given point, and so they do not even satisfy the Theorem of Jordan-Brouwer. On these grounds
the notion of a good pair arose and good pairs are the central topic of this article.

A solution for the points in the figure would be, to equip the black points with the 8-adjacency
and use the 4-adjacency for the white ones. Then is clear that a discrete notion of the 
Jordan-theorem is true for this example. 

For a long time adjacencies like the 4- and 8-neighborhood have been used, and of course, it 
is possible to generalize them to higher dimensions. This is done in this paper and we will
see, which pairs of such relations give us good pairs. To do so, we will use the gridcube model
of $\Z^n$ which is widely accepted and may be found in the book of A.~Rosenfeld and R.~Klette
\cite{klette}. It gives us a basic understanding of how these adjacencies may be build in high
dimensions and once we have a good mathematical description for them, we may use it for the study 
of pairs of the adjacencies that we will call ``cubical'' because of the relation to this 
model.

In the 1980s E.~Khalimsky \cite{khalimsky} proposed a topological motivated approach with
the so called Khalimsky-neighborhood. This topological notion gives also rise to graph-theoretic
adjacencies and so it seems interesting to study it. Since it is already known, that these
relations form good pairs, as seen in \cite{khalimsky} and \cite{herman}, we can use it as a
test for the theory that also shows, how we are able to combine topological and 
graph-theoretic concepts.

The paper is organized as follows: We start with some basic definitions in section 2 where
we do a tour through basic discrete topology and the 
graph-theoretic knowledge we use in this text, in section 3 the important concepts of
the paper \cite{paper01} are given and in section 4 we apply the theory to the aforementioned
adjacency relations. We end the text with some conclusions in section 5.

\section{Basic Definitions}\label{chapter_4}

\subsection{Topological Basics}

We use this section to introduce some basic topological notions. These stem from the usual 
set-theoretic topology as it might be found in any textbook on topology like
the one of St\"ocker and Zieschang \cite{sundz}, but we also introduce some facts 
given by P.S.~Alexandrov in his
text \cite{alexandroff}\footnote{Actually, Paul~Alexandroff is the same person as 
  Pavel~Sergeyevitch~Alexandrov. The different names origin in a different transcription of the 
  cyrillic letters in German and English.}.

\begin{Def}
  A pair $(\P,\TOP)$ is called \DF{topological space} for a set $\P$ 
  and set $\TOP\SUBSET\POT(\P)$, the so called \DF{open sets} or \DF{topology} on $\P$, 
  with the following properties: 
  \begin{enumerate}
  \item $O_i\in\TOP,i\in I \Rightarrow \bigcup_{i\in I}O_i \in \TOP$
  \item $O_1,\ldots,O_n\in\TOP \Rightarrow \bigcap_{1=1}^n O_i\in\TOP$
  \item $\P,\emptyset\in\TOP$
  \end{enumerate}
\end{Def}

A trivial topology on $\P$ is the \DF{discrete} topology
$\POT(\P)$. Please do not mistake the special ``discrete'' topology with the
``discrete'' setting we are working in. Even the $\R^n$ may be equipped with a discrete
topology and almost none of the discrete topologies we are referring to in this text are 
powersets of the base-set.

The subsets of $\P$, which have an open complement are called \DF{closed}. 
An open set $U\in\TOP$ is called \DF{neighborhood} of a point $x\in\P$ if 
$x$ is contained in $U$.

A topological space that satisfies the following stronger claim instead of 
property (2), is called \DF{Alexandrov-space}
\begin{enumerate}
\item[2'.] $O_i\in\TOP,i\in I$ $\THEN$ $\bigcap_{i\in I}O_i\in\TOP.$
\end{enumerate}
All results for topological spaces are also true in Alexandrov-spaces.
Topological spaces may be classified concerning the following separation
properties:
\begin{Def}
  A topological space may satisfy some of the separation axioms:
  \begin{description}
  \item[$T_0$:] $\FORALL_{x,y\in\P},x\neq y\EXISTS_{U\in\TOP}:(x\in
    U\not\ni y)\lor ( x\not\in U\ni y)$
  \item[$T_1$:] $\FORALL_{x,y\in\P},x\neq y\EXISTS_{U\in\TOP}:x\in
    U\not\ni y$
  \item[$T_2$:] $\FORALL_{x,y\in\P},x\neq y\EXISTS_{U,V\in\TOP}:(x\in
    U\not\ni y)\land ( x\not\in V\ni y)\land (U\cap V=\emptyset)$
  \end{description}
\end{Def}
One can see, that every $T_i$-space is also a $T_{i-1}$-space. It is also true,  
that considering property (2') interesing only for $T_0$-spaces:

\begin{Lemma}
  An Alexandrov-space that satisfies the separation axioms $T_1$ or $T_2$ necessarily 
  has the discrete topology.
\end{Lemma}

\begin{Beweis}
  Let $\P$ be a $T_1$-space, $p\in \P$ and $U$ a neighborhood of $p$. If 
  $U=\{p\}$, then we are done. Otherwise, there exists a $q\neq p$ in $U$
  and by property $(1)$, we may find a neighborhood $U'$, the contains $p$ but not $q$.
  The intersection of all these sets is open and so, $\P$ has to be discrete.
  
  The proof for $T_2$-spaces is analog.
\end{Beweis}

To give a topology on a set $\P$, it is enough to give a certain family $\mathcal{B}$
of open sets that can be used to generate all the open sets of $\P$ by using 
set-theoretic union. 
This family is then called \DF{base} of the topology $\TOP$.
A topological space is called \DF{locally finite}, if for any point $p$ in $\P$ exists a
finite open set and a finite closed set that both contain $p$.
In the following, we define how we can build new topological spaces from given ones.

\begin{Def}
  Let $(P_i,\TOP_i)$, $i\in I$, be a family of topological spaces and let
  $\P=\prod_{i\in I} \P_i$ be their product and $p_i:\P\TO \P_i$ projections. The
  \DF{product topology} $\TOP$ is defined by the base
  \[\mathcal{B}=\left\{\bigcap_{k\in K}p_k^{-1}(O_k):
  O_k\in\TOP_k,K\subset I, K\mbox{ finite}\right\}\enspace.\]
  The space $(\P,\TOP)$ is called \emph{topological product} of the
  $(\P_i,\TOP_i)$.
\end{Def}

\begin{Def}
  Let $(\P,\TOP)$ be a topological space and $A\SUBSET\P$. With the topology
  \[\TOP|_A=\ASET{O\cap A: O\in\TOP}\]
  The set $A$ can be turned into a topological space $(A,\TOP|_A)$. The topology $\TOP|_A$
  is called \DF{subspace topology} of $A$ with respect $\P$.
\end{Def}

\begin{Def}
  A mapping $f:\P\TO\mathcal{Q}$ between two topological spaces 
  $(\P,\TOP)$, $(\mathcal{Q},\TOPP)$ is called \DF{continuous}, if
  for every $O\in\TOPP$ the set $f^{-1}(O)$ is in $\TOP$.
\end{Def}

\begin{Def}
  A topological space $(X,\TOP)$ is called \DF{connected}, if it cannot be
  decomposed into two nonempty open sets:
  \[\P=O_1\cup O_2,\, O_1,O_2\in\TOP,\, O_1\neq\emptyset\neq O_2 \THEN
  O_1\cap O_2\neq\emptyset.\]
  A set $A\SUBSET \P$  is called connected, if it is connected in the subspace topology.
\end{Def}

\begin{Lemma}\label{stetigerzshg}
  Let $(\P,\TOP)$ be a connected topological space and let $(\mathcal{Q},\TOP')$ be a topological
  space. If $f:\P\TO\mathcal{Q}$ is a continuous mapping, then $\mathcal{Q}$  
  is connected.\qed
\end{Lemma}

%% \begin{Beweis}
%%   Assume for contradiction that $(\mathcal{Q},\TOP')$ is not connected.
%%   Then, sets $O_1,O_2\in\TOP'$ exist with 
%%   $O_1\neq\emptyset\neq O_2$, $O_1\cup O_2=Y$ and $O_1\cap
%%   O_2=\emptyset$. Therefore, $f^{-1}(O_1)$ and $f^{-1}(O_2)$ are open in
%%   $\P$. The preimages satisfy $f^{-1}(O_1)\neq\emptyset\neq f^{-1}(O_2)$ and $f^{-1}(O_1\cup
%%   O_2)=X$. Because of $f^{-1}(O_1\cap O_2)=\emptyset$, the set $\P$ would not be 
%%   connected.
%% \end{Beweis}

From the continuity of the projections $p_i$ in the definition of the product topology we can 
deduce the following

\begin{Lemma}\label{prodzshg}
  A topological space is connected if and only if all of its factors are connected. \qed
\end{Lemma}

We define a \DF{path} of length $m\in\N$ to be a continuous mapping
$w:\SET{0}{m}\TO \P$. A path is \DF{closed} if $w(0)=w(m)$.

\begin{Def}
  A topological space $(\P,\TOP)$ is called \DF{path-connected}, if for any two points
  $p,q\in \P$ exists a path $w$ of length $m$ depending only on $p$ and $q$, such 
  that $w(0)=p$ and $w(m)=q$.
  
  The topological space $(\P,\TOP)$ is called \DF{locally path-connected}
  if for every point $p\in\P$ and every neighborhood $U$ of $p$ a path-connected 
  neighborhood $V\SUBSET U$ exists.
\end{Def}

\begin{Folg} The following holds:
  \begin{enumerate}
  \item Path-connected spaces are connected.
  \item Connected and locally path-connected spaces are path-connected.\qed
  \end{enumerate}
\end{Folg}

\begin{Def}
  Let $X$ and $Y$ be topological spaces. A \emph{homotopy} from
  $X$ to $Y$ is a family of mappings $h_t:X\TO Y$, $t\in I=[0,1]$ 
  with the following property:
  The mapping $H:X\times I\TO Y$, $H(x,t)=h_t(x)$, is continuous. The set $X\times I$ 
  has the product topology.
  
  Two functions are called \DF{homotopic}, $f\cong g:X\TO Y$ if a homotopy $h_t:X\TO Y$
  exists with $h_0=f$ and $h_1=g$. If $g$ is constant then $f$ is called \emph{nullhomotopic}.

  A homotopy is called \DF{linear}, if it is linear in $t$.
\end{Def}

Just like in the definition of paths, the set $I$ does not need to be the set $[0,1]$ in the 
discrete setting we are going to use arbitrary connected subsets of $\Z$ for instance
$\SET{0}{m}\SUBSET\N$ with a fitting topology.

\begin{Def}
  A topological space is called simply connected if any closed path is nullhomotopic.
\end{Def}
This means that we continuously contract every closed path into one point.

\begin{Lemma}\label{einfachverein}
  If $(\P,\TOP)$ is a union of two open simply connected subspaces with contractible 
  intersection, then it is simply connected.\qed
\end{Lemma}

%%%%%%%%%%%%%%%%%%%%%%%%%%%%%%%%%%%%%%%%%%%%%%%%%%%%%%%%%%%%%%%%%%%%%%
%%
%% Alexandrov R\"aume
%%

\subsection{Alexandrov-Spaces}\label{section-alexandroff-space}

Every Alexandrov-space has an unique base that is given by the set of minimal 
neighborhoods of all points in the base-set. The minimal neighborhoods are easily
identified as the intersections of all neighborhoods of a given point. Let $p$ be a 
point in an Alexandrov-space $(\P,\TOP)$. We write $\U_\TOP(p)$ to denote its minimal
neighborhood. Analog we may find a minimal closed set containing a given point $p$. 
We denote this set by $C_\TOP(p)$. To create an analogy to the graph-theoretic background
of most of this theory, we define
\begin{equation}
  \A_\TOP(p):=(\U_\TOP(p)\cup\C_\TOP(p))\SETMINUS\{p\}
\end{equation}
to be the \DF{adjacency} of the point $p$ in $(\P,\TOP)$. The set $\A_\TOP(p)$ can be made 
to an Alexandrov-space in the subspace-topology.

Given a set $M\SUBSET\P$ we may analog define the sets:

\begin{equation}\label{eq:closure1}
  \U_\TOP(M) := \left\{p\in\P:\EXISTS_{q\in M}p\in\U_\T`OP(q) \right\}
\end{equation}

\begin{equation}\label{eq:closure2}
  \C_\TOP(M) := \left\{p\in\P:\EXISTS_{q\in M}p\in\C_\TOP(q) \right\}
\end{equation}

\begin{Lemma}\label{huellenop}
  The set functions $\U_\TOP$ and $\C_\TOP$ are closure operators, they satisfy:
  \begin{enumerate}
  \item $\U_\TOP(\emptyset)=\emptyset$.
  \item $M\SUBSET N\Rightarrow \U_\TOP(M)\SUBSET \U_\TOP(N)$.
  \item $\U_\TOP(\U_\TOP(M))=\U_\TOP(M)$.
  \end{enumerate}
\end{Lemma}

\begin{Beweis}
  The first property is trivial. To show the second one let $p\in\U_\TOP(M)$. 
  Therefore, it exists a $q\in M$ such that $p\in\U_\TOP(q)$.
  By the precondition we have $q\in N$ and therefore $p\in\U_\TOP(N)$. 

  To prove property 3, let $p\in\U_\TOP(\U_\TOP(M))$, therefore, a $q$ exists in $\U_\TOP(M)$
  such that $p\in\U_\TOP(q)$. If $q\in M$ holds, then holds $p\in\U_\TOP(M)$. 
  Otherwise, a $q'\in M$ exists such that $q$ is in $\U_\TOP(q')$.
  By the property $T_0$ of an Alexandrov-space, the point $p$ has to be in $\U_\TOP(q')$ 
  and therefore in $\U_\TOP(M)$. The other inclusion follows from 2.
\end{Beweis}

\begin{Lemma}\label{zus}
Let $(\P,\TOP)$ be an Alexandrov-space that contains one point $p$ such that the only open 
neighborhood of $p$ is the set $\P$ itself. Then $(\P,\TOP)$ is contractible.
\end{Lemma}

\begin{Beweis}
We define a homotopy $F:\P\times I\TO \P$ by $F(q,t)=q$
for $0\le t<1$ and $F(q,1)=p$ for each $q\in\P$. 

We show, that $F$ is continuos. Let $M\SUBSET \P$ be open.

Case 1: The point $p$ is in $M$. W.l.o.g. $M=\P$. Therefore, the set $F^{-1}(M)=\P\times I$
  is open.

Case 2: The point $p$ is not in $M$. The the set $F^{-1}(M)=M\times [0,1)$ is open.
\end{Beweis}

\begin{Lemma}\label{alexzus}
Let $(\P,\TOP)$ be an Alexandrov-space and $p\in\P$, then the set $\U_\TOP(p)$ is
contractible. Therefore, the Alexandrov-space $(\P,\TOP)$ has a base of contractible 
open sets. In particular, the set $(\P,\TOP)$ is local contractible.
\end{Lemma}

\begin{Beweis}
We utilize Lemma \ref{zus} together with $Y=\U(x)$ and $\omega=x$.
\end{Beweis}

It is possible to establish a notion of dimension in Alexandrov-spaces. It can also be
found in Evako et.al. \cite{evako}:

\begin{Def}\label{top_flaechen_def}
  Let $(\P,\TOP)$ be a Alexandrov-space and $p\in\P$.
  \begin{itemize}
  \item $\dim_\P(p) := 0$, if $\U_\TOP(p)\SETMINUS\{p\}=\emptyset$.
  \item $\dim(\P) :=n$, if there is a point $p$ in $\P$ such that $\dim_\P(p)=n$ 
    and for all $q\in\P$ exists a $k\le n$ with $\dim_\P(q)=k$.
 \item $\dim_\P(p) := n+1$, if $\dim(\U_\TOP(p)\SETMINUS\{p\})=n$. 
   The set $\U_\TOP(p)\SETMINUS\{p\}$ has the subspace topology.
  \item If no $k\in\N$ exists such that $\dim_\P(p)=k$ then define $\dim_\P(p)=\infty$. 
  \end{itemize}
\end{Def}

\begin{Def}
  We call $(\P,\TOP)$ a $0$-surface, if $\P$ has two points and is disconnected under 
  $\TOP$.
  
  The set $(\P,\TOP)$ is called $n$-surface for $n>0$, if $\P$ is connected under $\TOP$
  and for all $p\in\P$ the set $\A_\TOP(p)$ is a $(n-1)$-surface.

  A $n$-surface $(\P,\TOP)$ is called $n$-sphere, if $\P$ is finite and it is simply connected
  for $n>1$.
\end{Def}

By Evako et.al.\cite{evako} gilt:

\begin{Satz}\label{alexandroffflaeche}
  Let $(\P,\TOP)$ be a Alexandrov-space that is a $n$-surface for
  $n>2$. Then, for any point $p\in\P$ holds, that $\A_\TOP(p)$
  is simply connected. \qed
\end{Satz}

\begin{Satz}\label{alexandrov-partial-order}
  Every Alexandrov-space is a partial order and every partial order defines an 
  Alexandrov-space. \qed
\end{Satz}

%% \begin{Beweis}
%%   Use the base of the topology to define a partial order. And given a partial order use it 
%%   to define the topology.
%% \end{Beweis}

\subsection{The Khalimsky-Topology}\label{section-jordan-proof}

In this section we study an important Alexandrov-topologies. To define it we start with a topologization
of the set $\Z$ which we can interpret a a discrete line. What possibilities do we have to 
define a non-trivial topology on this set such that it is connected?

One can see, that the sets
\begin{equation}
\mathcal{B} = \{\{x\}: x\in\Z, x\equiv 0(2)\} \cup
\{\{x-1,x,x+1\}:x\in\Z,x\equiv 1(2)\}
\end{equation}
and 
\begin{equation}
\mathcal{B'} = \{\{x\}: x\in\Z, x\equiv 1(2)\} \cup
\{\{x-1,x,x+1\}:x\in\Z,x\equiv 0(2)\}
\end{equation}
are bases of topologies. They differ only by a translation. Therefore, it seams reasonable to
just choose one of them both. We will use the base $\mathcal{B}$ and denote its generated
topology by $\kappa$.

\begin{figure}[htb]
  \begin{center}
    \includegraphics{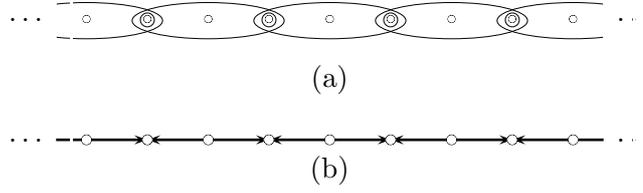}
  \end{center}
  \caption[Base of $(\Z,\kappa)$]{\small{Figure (a) shows a section of $(\Z,\kappa)$. 
      The base of the topology is represented by ellipses. The base of the topology may 
      also be depicted as a digraph. Figure (b) shows this.}}  
  \label{Pic:basistopologie}
\end{figure}

\begin{Lemma}
  The Alexandrov-space $(\Z,\kappa)$ is connected.\qed
\end{Lemma}

To go from here to the higher-dimensional case, we may view $\Z^n$ as a $n$-fold 
topological product of $\Z$. We denote the product topology 
with $\kappa_n$. By all we know so far, it is clear, that $(\Z^n,\kappa_n)$ is 
connected. We call this class of spaces \DF{Khalimsky-spaces} after E.~Khalimsky 
\cite{khalimsky}.

\begin{Lemma}
  The Alexandrov-space $(\Z^n,\kappa_n)$ is connected for all $n\ge 1$. 
\end{Lemma}

\begin{Beweis}
  This follows from Lemma \ref{prodzshg}.
\end{Beweis}

\begin{Satz}\label{svjb}
  All Khalimsky-spaces $(\Z^n,\kappa_n)$, $n\ge2$ satisfy the separation theorem of
  Jor\-dan-Brouwer.
\end{Satz}

\begin{Beweis}
  The proof is easy if one uses the methods of algebraic topology, because
  $(\Z^n,\kappa_n)$ is isomorphic to a cell-decomposition of $\R^n$:
  \begin{equation}
    \R^n =(\{\{i\}:i\in\Z,i\equiv 0(2)\}\cup\{(i-1,i+1):i\in\Z,i\equiv1(2)\})^n
  \end{equation}
  The set $(i-1,i+1)$ denotes the open real interval between the integers 
  $i-1$ and $i+1$.
  
  Since the Theorem of Jordan-Brouwer is true for any $\R^n$, $n\ge2$, it has to 
  hold for $n$-dimensional Khalimsky-space.

  We give another proof in section \ref{section-khalimsky-top}.
\end{Beweis}

%%%%%%%%%%%%%%%%%%%%%%%%%%%%%%%%%%%%%%%%%%%%%%%%%%%%%%%%%%%%%%%%%%%%%%
%%
%% Untersuchung konkreter Adjazenzen
%%

\subsection{Adjacency Relations}

To establish structure on the points of the set $\Z^n$ we have to
define some kind of connectivity relation. This might be done in terms of a
(set-theoretic) topology as in the last section, or we may develop a graph-theoretic 
framework as in the following part of the text.

\begin{Def}\label{adjazenzrelation}
  Given a set $\P$, a relation  $\alpha\SUBSET\P\times\P$ is called \DF{adjacency}
  if it has the following properties:
  \begin{enumerate}
  \item $\alpha$ is finitary: $\forall p\in\P:|\alpha(p)|<\infty$.
  \item $\P$ is connected under $\alpha$.
  \item Every finite subset  of $\P$ has at most one infinite connected component 
    as complement.
  \end{enumerate}
\end{Def}

A set $M\SUBSET\P$ is called \DF{connected} if for any two points 
$p,q$ in $M$ exist points $p_0,\ldots,p_m$ and a positive integer $m$ such that
$p_0=p$, $p_m=q$ and $p_{i+1}\in A(p_i)$ for all
$i\in\{0,\ldots,m-1\}$. Compare this definition to the topological one we gave above.

The property 3 of an adjacency-relation is in $\Z^n$ for $n\ge 2$ always satisfied.

In the text we will consider pairs $(\alpha,\beta)$ of adjacencies
on the set $\Z^n$. In this pair $\alpha$ represents the adjacency
on a set $M\SUBSET\Z^n$, while $\beta$ represents the adjacency on 
$\COMP{M}=\Z^n\SETMINUS M$.

Let $\T$ be the set of all translations on the set $\Z^n$.
The generators $\tau_1,\ldots,\tau_n\in\T$ of  $\Z^n$
induce a adjacency $\pi$ in a natural way: 

\begin{Def} Two points $p,q$ of $\Z^n$ are called \DF{proto-adjacent}, in terms
  $p\in\pi(q)$, if there exists a $i\in\{1,\ldots,n\}$ such that
  $p=\tau_i(q)$ or $p=\tau_i^{-1}(q)$. 
\end{Def}

We can view the generators of $\Z^n$ a the standard base of $\R^n$.

Another important adjacency on $\Z^n$ is \DF{$\omega$}.
\begin{equation}
  \omega(p):=\{q\in\Z: |p_i-q_i|\le 1, 0\le i\le n\}  
\end{equation}

%% \begin{Lemma}
%%   For every $n\ge 2$ and all $p\in\Z^n$ the set $\omega(p)$ is connected 
%%   under $\pi$. \qed
%% \end{Lemma}

In the rest of the text let $\alpha$ and $\beta$ be two adjacencies on 
$\Z^n$ such that for any $p\in\Z^n$ holds
\begin{equation}
  \pi(p)\SUBSET\alpha(p),\beta(p)\SUBSET\omega(p)\enspace.
\end{equation}

\begin{Lemma}
  The set $\Z^n$ is connected under $\pi$.\qed
\end{Lemma}

\section{Digital Manifolds}

If we want to talk about $(n-1)$-Manifolds in $\Z^n$ we have to give a proper definition. 
Unfortunatly, all the definitions known to the author from the literature are not usable
in terms of generalization to higher dimension or for the unification of the topological
and graph-theoretic approach. So it is necessary, to give a new definition that 
satisfies this two criteria. This is don in \cite{paper01}. The new definition is 
manly based on the so called \emph{separation property}. It gives a description on 
how a discrete $(n-1)$-manifold should look like locally. 

\subsection{The Separation Property}\label{Sec:trennungseigenschaft}

We call the set 
\begin{equation}
  C^k=\{0,1\}^k\times\{0\}^{n-k}\SUBSET\Z^n
\end{equation}
the $k$-dimensional standard \DF{cube} in $\Z^n$.
The set $C^k$ can be embedded in ${n}\choose{k}$ different ways in $C^n$.
A general $k$-cube in $\Z^n$ is defined by a translation of a standard cube.

Indeed, we can construct any $k$-cube $C$ from one point $p$ with $k$ generators
in the following way:
\begin{equation}
  C = \{\tau_1^{e_1}\cdot\tau_k^{e_k}(p): e_i\in\{0,1\}, i=1,\ldots,k \}
\end{equation}
The dimension of $C'$ is then $k+l$. We use this construction in the next definition.

\begin{Def}\label{trenndef}
  Let $M\SUBSET \Z^n$, $n \ge 2$ and $C$ be a $k$-cube,
  $2\le k\le n$. The complement of $M$ is in $C$ \DF{not separated} by $M$ under
  the pair $(\alpha,\beta)$, if 
  for every $\alpha$-component $M'$ of $C\cap M$ and every
  $(k-2)$-subcube $C^*$ of $C$ the following is true:
  
  If $C^*$ is such that $C^*\cap M'\neq\emptyset$ has maximal cardinality among 
  all sets of this form, and the sets $\tau_1(C^*)\SETMINUS M$ and 
  $\tau_2(C^*)\SETMINUS M$ are both nonempty and lie in one common $\beta$-component 
  of $\COMP{M}$, then holds
  \begin{equation}   
    (\tau_1\tau_2)^{-1}(\tau_1\tau_2(C^*)\cap M') \SUBSET
    \tau_1^{-1}(\tau_1(C^*)\cap M') \cap
    \tau_2^{-1}(\tau_2(C^*)\cap M')
    \enspace .
  \end{equation}
\end{Def}

\begin{figure}[htb]
  \begin{center}
    \includegraphics[scale=1.0]{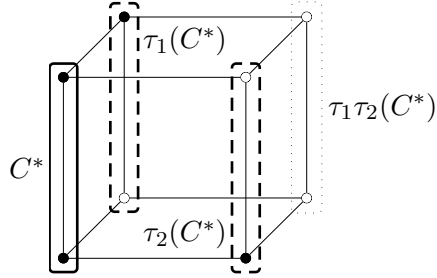}
  \end{center}
  \caption[The separation property]{\small{$C^*$ has an intersection
      of maximal cardinality with the $\alpha$-component
      $M'$. The sets $\tau_1(C^*)\SETMINUS M'$ and $\tau_2(C^*)\SETMINUS M'$ are
      nonempty and belong to a $\beta$-component of
      $\COMP{M}$. Since $\tau_1\tau_2(C^*)\cap M' = \emptyset$,
      the property of definition \ref{trenndef} is satisfied for this $C^*$.
      But the set $M'$ separates $\COMP{M}$ in the cube $C^*$. Why?}}  
  \label{Pic:trennungseigenschaft1}
\end{figure}

In the following, we only consider the case when $C\cap M$ has at most one
$\alpha$-component. This can be justified by viewing any other
$\alpha$-component besides the one considered as part of the
background, since there is no $\alpha$-connection anyway.
This property also gets important if we study the construction of the
simplicial complex.

A set $M$ has the \DF{separation property} under a pair $(\alpha,\beta)$, 
if for every $k$-cube $C$, $2\le k\le n$ as in the definition \ref{trenndef} 
the set $\COMP{M}$ is in $C$ not separated by $M$

\begin{figure}[htb]
  \begin{center}
    \includegraphics[scale=1.0]{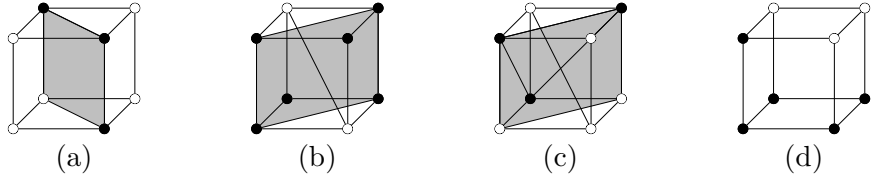}
  \end{center}
  \caption[The separation property 2]{\small{The black points represent the set $M$ in the
      given 3-cubes. The white points represent the complement of $M$. In the cases (a) to (c)
      the complement, which is connected, is separated by $M$. This separation is depicted 
      by the gray plane spanned by $C^*$ and $\tau_1\tau_2(C^*)$. 
      In Figure (d) occurs no separation, 
      since the only choice for $C^*$ would be a 1-cube, that contains only black 
      points. 
      %In one case the relation claimed by the definition \ref{trenndef}
      %in the other cases the sets $\tau_1(C^*)\SETMINUS M'$ respectively 
      %$\tau_2(C^*)\SETMINUS M'$ are empty. Therefore, the postulation of the 
      %definition is trivially true.
  }} 
  \label{Pic:trennungseigenschaft}
\end{figure}

The meaning of the separation property is depicted in the figure
\ref{Pic:trennungseigenschaft}.

\begin{Def}\label{n-1-mannigfaltigkeit}
  An $\alpha$-connected set $M\SUBSET\Z^n$, for $n\ge 2$, is a
  \DF{(digital) $(n-1)$-manifold} under the pair
  $(\alpha,\beta)$, if the following properties hold:
  \begin{enumerate}
  \item\label{cubeconnection} In any $n$-cube $C$ the set $C\cap M$
    is $\alpha$-connected.
  \item\label{twocomponents} For every $p\in M$ the set
    $\omega(p)\SETMINUS M$ has exactly two $\beta$-components
    $C_p$ and $D_p$. 
  \item\label{componentunity} For every $p\in M$ and every
    $q\in\alpha(p)\cap M$ the point $q$ is $\beta$-adjacent to $C_p$ and $D_p$.
  \item\label{separationproperty} $M$ has the separation property.
  \end{enumerate}
\end{Def}

How should a $(n-1)$-manifold look like globally in general? We do not know. But we might say,
that a single point in $\Z^n$ might be considered as the inside of some object, i.e. 
that it might be separated by the other points. The way to do this is to require the set
of neighbors of a point to be a $(n-1)$-manifold. This justifies the following:

\begin{Def}
A pair $(\alpha,\beta)$ of adjacency relations on $\Z^n$ is a
\DF{separating pair} if for all $p\in\Z^n$ the set $\beta(p)$ is a
$(n-1)$-manifold under $\alpha$.
\end{Def}

\subsection{Double Points}
\begin{Def}\label{doppelpunkt}
  A point $p\in\beta(z)$ $z\in\Z^n$ is a \DF{double point} under the
  pair $(\alpha,\beta)$, if there exist points
  $q\in\pi(z)\cap\alpha(p)$ and $r\in\beta(z)\cap\pi(p)$ 
  and a simple\footnote{A translation $\tau$  is called \DF{simple} if no other translation
  $\sigma$ exists with $\sigma^n=\tau$, $n\in\Z, |n|\neq 1$. } 
  translation $\tau\in\T$ with $\tau(p)=q$, $\tau(r)=z$ and $q\in\alpha(r)$. 
\end{Def}

This concept is the key to a local characterization of the good pairs 
$(\alpha,\beta)$. Without it, one could not consistently define topological
invariants like the Euler-char\-acter\-istic. It means that an edge
between points in a set $M$ can be crossed by an edge between points
of its complement and these four points lie in a square defined by the
corresponding adjacencies. This crossing can be seen as a double point,
belonging both to the foreground \emph{and} to the background. Also,
mention the close relationship to the separation property, which is a
more general concept of similar interpretation. For further insight, refer 
to the text \cite{paper01}.

\begin{figure}[htb]
\begin{center}
\includegraphics{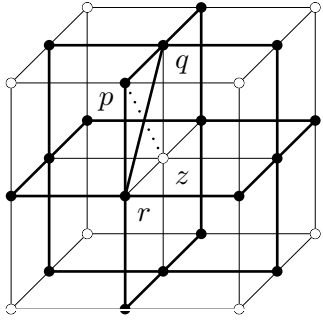}
\end{center}
\caption[A double point]{\small{A double point $p$. The fat edges
    represent the $\alpha$-adjacency. Only the relevant edges have been drawn for
    clarity. The dotted edge represents the $\beta$-adjacency of $p$ and $z$. The 
    black points are the $\beta$-neighbors of $z$.}} 
\label{Pic:doppelpunkt}
\end{figure}

\begin{Def}\label{goodpair}
  A separating pair of adjacencies $(\alpha,\beta)$ in
  $\Z^n$ is a \DF{good pair}, if for every $p\in\Z^n$ the set
  $\beta(p)$ contains no double points. 
\end{Def}

\section{Good Pairs of Adjacency Relations}\label{chapter_6}

%%%%%%%%%%%%%%%%%%%%%%%%%%%%%%%%%%%%%%%%%%%%%%%%%%%%%%%%%%%%%%%%%%%%%%
%%
%%  kubische Adjazenzen
%%

\subsection{Cubical Adjacencies}\label{kubische_adjazenzen}

We will study adjacencies in the sense of the gridcube-model. This is a common model
in computer graphics literature and has nothing to do with the $n$-cubes we talked about
earlier. We use this model here to make it easy to study the adjacency relations in this 
section. For more on this topic refer to the Book of Rosenfeld and Klette \cite{klette}
 
We identify the points of $\Z^n$ with $n$-dimensional unit-cubes with barycenters in the
points of the lattice $\Z^n$. The cube $W$ that represents the point $0\in\Z^n$ can be
expressed in euclidean space as $[-\frac{1}{2},\frac{1}{2}]^n$. Those gridcubes may
be interpreted as union of (polytopal) faces of different dimension. Any of its faces 
is again a gridcube, only with a lower dimension. Take, for instance, a 3-dimensional
gridcube $[-\frac{1}{2},\frac{1}{2}]^3$. It has, among others, the 0-dimensional face 
$(\frac{1}{2},\frac{1}{2})$, the 1-dimensional face 
$[(-\frac{1}{2},\frac{1}{2}),(\frac{1}{2},\frac{1}{2})]$
and the 2-dimenional face with the vertices
$(-\frac{1}{2},\frac{1}{2}), (\frac{1}{2},\frac{1}{2}), 
(-\frac{1}{2},-\frac{1}{2})$ and $(\frac{1}{2},-\frac{1}{2})$.

Two given gridcubes may share a $k$-dimensional face for $0\le k<n$. This
$k$-face is just the intersection of both of them. So we might say that the elements
$(0,\ldots,0)$ and $(1,\ldots,1)$ of $\Z^n$ intersect in a common vertex (0-face) with
the coordinates $(\frac{1}{2},\ldots\frac{1}{2})$. However, the elements $(0,\ldots,0)$ 
and $(1,0,\ldots,0)$ share a common $(n-1)$-face.

In the rest of the text we will no longer make the gridcube model explicit. It just serves
as an introduction to visualize the concepts that we use to analyze the discrete
geometry even in higher dimensions\footnote{I find it a lot easier to imagine a four-dimensional 
cube, than a four-dimensional grid...}.

\begin{Def} Two points $p,q\in\Z^n$ are called \DF{$k$-adjacent} for $0\le
  k<n$, denoted by $p\in\alpha_k(q)$, if their corresponding gridcubes share a common
  $k$-face. We call this adjacencies \DF{cubical}.
\end{Def} 

Clearly, this kind of relation we just defined is an adjacency-relation in the sense of 
definition \ref{adjazenzrelation}:

\begin{Lemma}
  The relation $\alpha_k$ is an adjacency-relation on $\Z^n$ 
  for every $n\ge 2$ and all integers $k$ between $0$ and $n-1$.
\end{Lemma}

\begin{Beweis}
  First, we have to check that for any $p\in\Z^n$ the set $\alpha_k(p)$ has only
  finite cardinality. It is easy to check, that $\alpha_0(p)$ is just $\omega(p)$ 
  as defined earlier and every $\alpha_k(p)$ for $0\le k<n$ is a subset of $\omega(p)$.
  Since $\omega(p)$ has $3^n-1$ Elements in $\Z^n$, the relations $\alpha_k$ must be 
  finitary.

  To see that $\Z^n$ is connected under any $\alpha_k$, $0\le k<n$, we observe that 
  $\alpha_{n-1}$ is just another interpretation for the relation $\pi$ defined earlier.
  Since $\Z^n$ is $\pi$-connected as proven in \cite{paper01} and every $\alpha_k$ is a 
  superset of $\alpha_{n-1}$, we conclude that $\Z^n$ is $\alpha_k$-connected.

  The last property is in $\Z^n$ with $n\ge 2$ trivially satisfied.
\end{Beweis}

\begin{Lemma}\label{kuben_representation}
  The cubical adjacency $\alpha_k(x_1,\ldots,x_n)$ may be represented in $\Z^n$ as the 
  set:
  \begin{equation}
    \left\{(y_1,\ldots,y_n)\in\Z^n: 
    \max_{i=1,\ldots,n}\{|x_i-y_i|\}=1,1\le\sum_{i=1}^n|x_i-y_i|\le n-k\right\}
  \end{equation}
\end{Lemma}

\begin{Beweis}
  Let $p$ and $q$ be two points of $\Z^n$ such that $p\in\alpha_k(q)$. 
  This means, the gridcubes corresponding to $p$ and $q$ share a common $k$-face.
  Their distance in the maximum-metric may not be greater than 1. Furthermore,
  $p$ and $q$ may not share a single common $l$-face for $0\le l<k$. That means, all of that 
  $l$-faces must be faces of common $k$-faces. Therefore, the two points may not have
  more than $k$ coordinates in common.
\end{Beweis}

\begin{Lemma}
Let $\alpha$ be a cubical adjacency on $\Z^n$. It holds:
\begin{enumerate}
\item $\alpha$ is invariant under translations
\item $\alpha$ is invariant under permutations of coordinates.
\end{enumerate}
\end{Lemma}

\begin{Beweis}
Let $\tau$ be any translation on $\Z^n$. We need to show $\tau(\alpha(p))=\alpha(\tau(p))$ 
for any $p\in\Z^n$. 
From the representation of $\alpha(p)$ we may deduce:
\begin{eqnarray}
\tau(\alpha(p)) &=& \tau(\{q\in\Z^n:q\in\alpha(p)\} \\
                &=& \{\tau(q):q\in\Z^n, q\in\alpha(p)\} \\
                &=& \{q'\in\Z^n: q'\in\alpha(\tau(p)) \} \\
                &=& \alpha(\tau(p))
\end{eqnarray}
The proof of the second part is analog.
\end{Beweis}

What is the structure of the cubical adjacencies in $\Z^n$? We take a closer look
at $n$-dimensional cubes.

\begin{Lemma}
  The number of $k$-faces of a $n$-dimensional cube is
  \begin{equation}
    {{n}\choose{k}}\cdot 2^{n-k}\enspace.
  \end{equation}
\end{Lemma}

\begin{Beweis}
  We use induction on the dimension $n$ of the cube.

  For $n=0$ we observe, that a 0-dimensional cube is just a point and has only one 0-face.
  Therefore, the induction base is correct.

  In the case $n>0$, we notice that a $n$-dimensional cube may be created from a 
  $(n-1)$-dimensional one by doubling the cube and inserting a $k$-face for every 
  $(k-1)$-face in the original cube. Therefore, we get by induction hypothesis and Pascals
  Theorem:
  \begin{eqnarray}
    2{{n-1}\choose{k}}\cdot2^{n-1-k}+{{n-1}\choose{k-1}}\cdot2^{n-1-(k-1)}
    &=& \left[{{n-1}\choose{k}}+{{n-1}\choose{k-1}}\right]\cdot2^{n-k} \\
    &=& {n \choose k}\cdot2^{n-k}
  \end{eqnarray}
  This proves the Lemma.
\end{Beweis}

\begin{Lemma}
  For every $p\in\Z^n$, the number of $k$-neighbors is
  \begin{equation}
    |\alpha_k(p)| = \sum_{i=k}^n{{n}\choose{i}}\cdot2^{n-i}\enspace.
  \end{equation}
\end{Lemma}

\begin{Beweis}
  Obviously, any $l$-face $\sigma$ of a cube contains at least one $k$-face
  $\tau$ for $0\le k\le l \le n$. Therefore, $k$-adjacent cubes exist, that are also $l$-adjacent. 
  Since that are those, that share more than one common $k$-face, the set
  $\alpha_k(p)$ for $p\in\Z^n$ may be decomposed into the following disjoint sets:
  \begin{equation}
    \begin{split}
      \alpha_k(p) &=\{q\in\Z^n:\mbox{$p,q$ have at most one $k$-face in common }\} \\
      &\cup\,\,\{q\in\Z^n:\mbox{$p,q$ have at most one $(k+1)$-face in common }\}\\ 
      &\,\,\,\,\vdots \\
      &\cup\,\,\{q\in\Z^n:\mbox{$p,q$ have at most one $(n-1)$-face in common }\} 
    \end{split}
  \end{equation}
  By adding the cardinalities of these sets, which we can easily compute with
  the last Lemma we get the result
  $\alpha_k(p)=\sum_{i=1}^n{{n}\choose{i}}2^{n-i}$.
  This proves the Lemma.
\end{Beweis}

By this technique we get as examples of cubical adjacencies in $\Z^2$ the known
4- and 8-adjacencies, in $\Z^3$ the 6-, 18- and 26-adjacencies and in $\Z^4$ the
8-, 32-, 64- and 80-adjacencies.

%%%%%%%%%%%%%%%%%%%%%%%%%%%%%%%%%%%%%%%%%%%%%%%%%%%%%%%%%%%%%%%%%%%%%%
%%
%% Gute Paare regul\"arer Adjazenzen
%%

\subsection{Good Pairs of Cubical Adjacencies}

In this section we will study, how we have to choose two cubical adjacencies to get to a
good pair. We first will see, that it does not matter at which point of $\Z^n$ we study
the adjacency, since the neighborhoods of all points look the same. 

\begin{Lemma}
  Let $\alpha$ be a cubical adjacency in $\Z^n$. For any $p\in\Z^n$ the set
  $\alpha(p)$ is graph-theoretical isomorphic to $\alpha(0)$.
\end{Lemma}

\begin{Beweis}
  This follows from the invariance under translations and the symmetry of the cubical 
  adjacencies.
\end{Beweis}

\begin{Lemma}\label{kleiner-zushang}
  Let $M\SUBSET\Z^n$ be $\alpha_k$-connected. Then $M$ is also
  $\alpha_l$-connected for $0\le l<k\le n-1$.
\end{Lemma}

\begin{Beweis}
  Let $M$ be $\alpha_k$-connected. Thus, we have for any two $p,q\in M$ a path 
  $p=p^{(0)},\ldots,p^{(a)}=q$ such that $p^{(i)}\in M$ for
  $i\in\{0,\ldots,a\}$ and $p^{(i-1)}\in\alpha_k(p^{(i)})$ for
  $i\in\{1,\ldots,a\}$. By definition of $\alpha_k$, Lemma \ref{kuben_representation} and
  $l<k$ holds for $p^{(i-1)}$ and $p^{(i)}$:
  $|p^{(i)}_j-p^{(i-1)}_j|\le 1$ and 
  \begin{equation}
    1\le\sum_{j=1}^n|p^{(i)}_j-p^{(i-1)}_j|\le n-k < n-l\enspace.
  \end{equation}
  Therefore, we have $p^{(i-1)}\in\alpha_k(p^{(i)})$ for $i\in\{1,\ldots,a\}$
  and the path $p=p^{(0)},\ldots,p^{(a)}=q$ is also a $\alpha_l$-path.
\end{Beweis}

The next Lemmata help us understand, which adjacencies may be used as good pair.

\begin{Lemma}
  Let $(\alpha_l,\alpha_k)$ be a pair of cubical adjacencies on $\Z^n$,
  $n\ge 2$. For any $n$-cube $C$ as in section \ref{Sec:trennungseigenschaft}, the set
  $C\cap\alpha_k(0)$ is connected under $\alpha_l$ if the following holds:
  \begin{enumerate}
  \item $0\le k\le n-2$ and $0\le l\le n-1$, or
  \item $k=n-1$ and $0\le l\le n-2$.
  \end{enumerate}
\end{Lemma}

\begin{Beweis}
  1. We use Lemma \ref{kleiner-zushang} and prove the proposition for $l=n-1$

  Let $C'$ be any subcube of $C$, that does not contain the point 0.
  We first show that $C'\cap M$ is $\alpha_l$-connected. Suppose w.l.o.g. that
  the point $p=(1,0,\ldots,0)$ is in $C'$ and choose any other point $r\in C'\SETMINUS M$.
  The point $r$ then has the form $r=(1,r_2,\ldots,r_n)$ with
  \begin{equation}
    \max_{i=1,\ldots,n}|r_i|=1 \mbox{ and } 1\le 1 +
    \sum_{i=2}^n|r_i|\le n-k\enspace.
  \end{equation}
  We select the smallest index $i\in\{2,\ldots, n\}$ such that $r_i\neq 0$ and define 
  \begin{equation}
    r'=(1,r_2,\ldots,r_{i-1},0,r_{i+1},\ldots,r_n)\enspace.
  \end{equation}
  The point $r'$ is in $\alpha_{n-1}(r)$:
  \begin{equation}
    \max_{i=1,\ldots,n}|r'_i|=1 \mbox{ and } 1\le
    \sum_{i=1}^n|r'_i-r_i|=|r'_i-r_i|=1\le n-(n-1)\enspace.
  \end{equation}
  By iterating this process we get an $\alpha_{n-1}$-path from $r$ to $p$.

  Let now be $C'$ and $C''$ be two different $(n-1)$-cubes. We may suppose w.l.o.g.
  that $p=(1,0,\ldots,0)\in C'$ and $q=(0,1,0,\ldots,0)\in C''$.
  The two cubes contain a common point $t=(1,1,0,\ldots,0)$ in $M$ since this point is
  $\alpha_{n-1}$-adjacent to $p$ and $q$ and it is in $\alpha_k(0)$ for 
$0\le k\le n-2$:
  \begin{equation}
    \max_{i=1,\ldots,n}|t_i|=1 \mbox{ and } \sum_{i=1}^n|t_i| = 2\le n-k\enspace.
  \end{equation}
  Therefore, the set $C\cap\alpha_k(0)$ is $\alpha_{n-1}$-connected.

  2. We show, that $C\cap\alpha_{n-1}$ is  connected under $\alpha_{n-2}$.
  By Lemma \ref{kleiner-zushang} this is enough.

  The set $C\cap\alpha_{n-1}$ contains all points $p^{(i)}=(p_1,\ldots,p_n)$, such that
  exactly one $i\in\{1,\ldots,n\}$ exists with $p^{(i)}_i\neq 0$ and 
  $|p^{(i)}_i|=1$. Let $p^{(i)}$ and $p^{(j)}$ be two such points with $i\neq j$.
  We have 
  \begin{equation}
    \max_{m=1,\ldots,n}|p^{(i)}_m-p^{(j)}_m|=1 \mbox{ and }
  \sum_{m=1}^n|p^{(i)}_m-p^{(j)}_m|=2\le n-(n-2)\enspace. 
  \end{equation}
  Therefore, $p^{(i)}$ and $p^{(j)}$ are $\alpha_{n-2}$-connected.
\end{Beweis}

\begin{Folg}\label{kub-surf}
  Given a pair $(\alpha_l,\alpha_k)$, then $\alpha_k(0)$ is
  $\alpha_l$-connected, if the following holds:
  \begin{enumerate}
  \item $0\le k\le n-2$ and $0\le l\le n-1$ or
  \item $k=n-1$ and $0\le l\le n-2$.
  \end{enumerate}
\end{Folg}

\begin{Beweis}
  This follows from the configuration of the $n$-cubes in $\omega(0)$ and the distribution
  of the $\pi$-neighbors of 0 in those $n$-cubes
\end{Beweis}

\begin{Lemma}
  Let $(\alpha_l,\alpha_k)$ be a pair of cubical adjacencies on $\Z^n$ with $n\ge 2$. 
  Then the set $\omega(p)\SETMINUS\alpha_k(0)$ has exactly two $\alpha_k$-components
  for all $p\in\alpha_{k}{0}$.
\end{Lemma}

\begin{Beweis}
  Obviously, 0 is in $\omega(p)$ for any $p\in\alpha_k(0)$ and it has no other 
  $\alpha_k$-neighbors in $\alpha_k(0)\SETMINUS\omega(p)$.

  We choose any point $p$ in $\alpha_k(0)$. Then, $\omega(p)$ contains points $s$ with
  $\max_{i=1,\ldots,n}|s_i|=2$. Those are not contained in in $\alpha_k(0)$ and form a
  $\pi$-connected set. Therefore they are also $\alpha_k$-connected.

  Define the set:
  \begin{equation}
    \omega(p)_i:=\{s\in\omega(p): |s_i|=2\}\enspace.
  \end{equation}

  W.l.o.g. we consider $\omega(p)_1$ that contains the point $p'=(2,p_2,\ldots,p_n)$. 
  It is easy to see, that either the point $p''=(-2,p_2,\ldots,p_n)$ or the point
  $p'$ is in $\omega(p)_1$.

  Let $s=(2,s_2,\ldots,s_n)$ be any point in $\omega(p)_i$. 
  We construct a $\pi$-path from $s$ to $p'$ by defining the point
  \begin{equation}
    s'=(2,\ldots,p_2,\ldots,p_{i-1},p_i,s_{i+1},\ldots,s_n)
  \end{equation}
  with the smallest index $i\in\{2,\ldots,n\}$ such that $s_i\neq p_i$.
  The point $s'$ is a $\pi$-neighbor of $s$ and after a finite number of iterations 
  we have the $\pi$-path from $s$ to $p'$.
  The sets $\omega(p)_i$ and $\omega(p)_j$ contain the points
  \begin{equation}
    (p_1,\ldots,p_{i-1},2,p_{i+1},\ldots,p_n)
  \end{equation}
  and
  \begin{equation}
    (p_1,\ldots,p_{j-1},2,p_{j+1},\ldots,p_n)\enspace,
  \end{equation}
  respectively. In both sets the point
  \begin{equation}
    t=(p_1,\ldots,p_{i-1},2,p_{i+1},\ldots,p_{j-1},2,p_{j+1},\ldots,p_n)
  \end{equation}
  is contained and therefore, the sets are $\pi$-connected.

  It remains to show, that points in $\omega(0)\SETMINUS\alpha_k(0)\cap\omega(p)$
  are $\pi$-adjacent to one of the $\omega(p)_i$. Let $i\in\{1,\ldots,n\}$, such that
  $s_i=p_i\neq 0$. 
  In the case $p_i>0$, then we have
  \begin{equation}
    s'=(s_0,\ldots,s_{i-1},s_i+1,s_{i+1},\ldots,s_n)\in\omega(p)_i
  \end{equation}
  and in the case $p_i<0$, it holds
  \begin{equation}
    s'=(s_0,\ldots,s_{i-1},s_i-1,s_{i+1},\ldots,s_n)\in \omega(p)_i\enspace.
  \end{equation}

  Finally, we have to observe the case of the point $s$ with $0=s_i\neq p_i$. 
  Then, the point
  \begin{equation}
    s'=(s_1,\ldots,s_{i-1},p_i,s_{i+1}\ldots,s_n)
  \end{equation}
  is a $\pi$-neighbor
  of $s$, that is not in $\alpha_k(0)$.
  This follows from
  \begin{equation}
    n-k<\sum_{j=1}^n|s_j|<\sum_{j=1}^{i-1}|s_j|+|p_i|+\sum_{j+1}^n|s_j|
    =\sum_{j=1}^n|s_j|+1\enspace.
  \end{equation}
  Thus, in the set $\alpha_k(0)\SETMINUS\omega(p)$, there is only one $\alpha_k$-component
  different from 0.
\end{Beweis}

\begin{Lemma}\label{zahl-der-komponenten-von-alpha-k}
  Let $(\alpha_l,\alpha_k)$ be a pair of cubical adjacencies on
  $\Z^n$ with $n\ge 2$. For any $p\in\alpha_k(0)$ any point
  $q\in\alpha_l(p)\cup\alpha_k(0)$ is $\alpha_k$-adjacent to both of the
  $\alpha_k$-components of $\omega(p)\SETMINUS\alpha_k(0)$, if the following 
  holds:
  \begin{enumerate}
  \item $1\le k\le n-1$ and $0\le l\le n-1$
  \item $0\le k\le n-2$ and $l=n-1$
  \end{enumerate}
\end{Lemma}

\begin{Beweis}
  1. Let $p\in\alpha_k(0)$ and $q\in\alpha_l\cap\alpha_k(0)$ be arbitrary chosen.
  Since $q\in\alpha_k(0)$, the point $q$ is $\alpha_k$-adjacent to the
  $\alpha_k$-component $\{0\}$.

  We define the set
  \begin{equation}
  I(p,q):=\{1\le i\le n:p_i=q_i=0\}\enspace.
  \end{equation}
  This set is non-empty since $1\le k\le n-1$. Let $q'=(q'_1,\ldots,q'_n)$ be the point with
  the following coordinates 

  \begin{equation}
  q'_i=\left\{\begin{array}{cl}
  q_i&\mbox{if }q_i\neq 0\\
  p_i&\mbox{if }q_i=0\mbox{ and }p_i\neq 0\\
  \end{array}\right.
  \end{equation}
  The remaining $q'_i$ for $i\in I(p,q)$ will be assigned with the values $\pm1$ and 0
  such that exactly $n-k+1$ of $n$ coordinates are different from 0.

  The point $q'$ is not contained in $\alpha_k(0)$, since
  \begin{equation}
    \sum_{i=1}^n|q_i'|=n-k+1>n-k\enspace.
  \end{equation}
  Because of $|q_i'-p_i|\le 1$, the point $q'$ must be in $\omega(p)$. Therefore, the point
  $q'$ is in $\omega(p)\SETMINUS(\alpha_k(0)\cup\{0\})$.
  Since $q$ is not 0, we have
  \begin{equation}
    \sum_{i=1}^n|q_i'-q_i|=\sum_{i=1}^n|q_i'|-\sum_{i=1}^n|q_i|\le
    n-k\enspace.
  \end{equation}
  Therefore, the point $q'$ is in $\alpha_k(q)$ and the set 
  $\omega(p)\SETMINUS(\alpha_k(0)\cup\{0\})$ is
  $\alpha_k$-adjacent to $q$.

  For $k=n-1$ and $l=n-1$ the set $\alpha_l(p)\cap\alpha_k(0)$ is empty.
  Thus the proposition is true.

  We may not choose $k$ as 0 for $0\le l\le n-2$, as the following example shows:
  Let $p\in\pi(0)$ and
  $q\in\alpha_l(p)\cap\pi(0)$, then
  \begin{equation}
    \sum_{i=1}^n|p_i-q_i|=2\le n-l\enspace,
  \end{equation}
  and thus
  $q\in\alpha_l(p)$. The point $q$ has no $\alpha_0$-neighbors
  in $\omega(p)\SETMINUS\alpha_0(0)$, because of $\alpha_0=\omega$ and
  because $r_i=\pm2$ for $p_i=\pm1$, hold for all $r\in\omega(p)\SETMINUS(\omega(0)\cup\{0\}$. 

  2. Let $p$ be a point in $\alpha_k(0)$ and let
  $q$ be any point in $\alpha_{n-1}(p)\cap\alpha_k(0)$. Obviously, the point 
  $q$ is $\alpha_k$-adjacent to $\{0\}$.

  We choose
  \begin{equation}
    i\in\{1\le i\le n: p_i\neq 0\mbox{ and }
    q_i\neq 0\}\enspace.
  \end{equation}
  This is a non-empty set, because the points $q$ and $p$ coincide in at least one
  non-zero coordinate, since $q\in\alpha_{n-1}(p)\cap\gamma_k(0)$.
  We define
  \begin{equation}
    q'=\left\{\begin{array}{cl}
    (q_1,\ldots,q_{i-1},q_i+1,q_{i+1},\ldots,q_n)&\mbox{if } p_i=1\\
    (q_1,\ldots,q_{i-1},q_i-1,q_{i+1},\ldots,q_n)&\mbox{if } p_i=-1\\
    \end{array}\right.
  \end{equation}
  We have $|q_j'-p_j|\le 1$ for all $i\in\{1,\ldots,n\}$. Therefore, the point
  $q'$ is in $\omega(p)$. Thus, $|q_i|=2$, otherwise $|q_i-p_i|$ would not be 
  smaller or equal to 1.
  We conclude that $q'$ is no point in $\alpha_k(0)$ and $q'\neq 0$.
  The point $q'$ is therefore a member of $\omega(p)\SETMINUS(\alpha_k(0)\cup\{0\})$ 
  and it is a $\pi$-neighbor of $q$. Which means it is an
  $\alpha_k$-neighbor, too.
  Thus, the point $q$ is $\alpha_k$-adjacent to
  $\omega(p)\SETMINUS(\alpha_k(0)\cup\{0\})$. 
\end{Beweis}

\begin{Lemma}
  Let $(\alpha_l,\alpha_k)$ be a pair of cubical adjacencies. Then the set
  $\alpha_k(0)$ satisfies the separation property under this pair.
\end{Lemma}

\begin{Beweis}
  Instead of $\alpha_k(0)$, we consider the set 
  $\overline\alpha_k(0)=\alpha_k(0)\cup\{0\}$. We may do so, because the point $0$
  is a different $\alpha_k$-component of $\COMP{\alpha_k(0)}$. It is not separable
  and therefore has no influence of the separability of the other points. If we know
  whether $\overline\alpha(0)$ has the separation property, 
  then we also know that $\alpha_k(0)$ has it too.

  Let $C$ be a $m$-cube, $2\le m\le n$, containing a point of $\overline\alpha_k(0)$ and 
  let $C^*$ be a $(m-2)$-subcube of $C$ that has a maximal intersection with
  $\overline\alpha_k(0)$. There exist two translations $\tau_1$ and $\tau_2$ such that we may 
  decompose $C$ in the following way:
  \begin{equation}
    C=C^*\cup\tau_1(C^*)\cup\tau_2(C^*)\cup\tau_1\tau_2(C^*)\enspace.
  \end{equation}

  Case 1: The translated cubes $\tau_i(C^*)$, $i=1,2$ and $\tau_1\tau_2(C^*)$ each 
  contain a point $p$ such that $\max_{i=1,\ldots,n}|p_i|=2$.
  Then, one can easily see that $\tau_1\tau_2(C^*)$ is fully in $\COMP{M}$.
  And so we have
  \begin{equation}
    (\tau_1\tau_2)^{-1}(\tau_1\tau_2(C^*)\cap M)\SUBSET
    \tau_1^{-1}(\tau_1(C^*)\cap M)\cap\tau_2^{-1}(\tau_2(C^*)\cap M)
    \SUBSET C^*\cap M\enspace.
  \end{equation}
  This is true, especially if
  $\tau_i(C^*)\SETMINUS\overline\alpha_k(0)\neq\emptyset$, $i=1,2$ and both of the 
  set are $\alpha_k$-connected.

  Case 2: The cube $C$ contains no point $p$ such that 
  $\max_{i=1,\ldots,n}|p_i|=2$. Let $q$ be the point in $C^*$  that satisfies
  $\sum_{i=1}^n|q_i|=x$ and $0\le x\le n-k$ be minimal in $C$. 
  It is sufficient to claim this minimality as the following consideration shows:
  We have:
  \begin{equation}
    C^*\cap M=\left\{p:\sum_{i=1}^n|p-i|\le \min(m-2,n-k)-x \right\}
  \end{equation}
  \begin{equation}
    \tau_{1,2}C^*\cap M=\left\{p:\sum_{i=1}^n|p-i|\le
    \min(m-3,n-k-1)-x \right\}
  \end{equation}
  \begin{equation}
    \tau_1\tau_2(C^*)\cap M=\left\{p:\sum_{i=1}^n|p-i|\le
    \min(m-4,n-k-2)-x \right\}
  \end{equation}
  The cube $C^*$ has always a maximal number of points in $\overline\alpha_k(0)$.
  If $\tau_{1,2}(C^*)$ and $\tau_1\tau_2(C^*)$, respectively contain a maximal number of
  points in $\overline\alpha_k(0)$, so they are both contained in
  $\overline\alpha_k(0)$.

  Therefore, we have the following inclusions:
  \begin{equation}
    (\tau_1\tau_2)^{-1}(\tau_1\tau_2(C^*)\cap M)\SUBSET
    \tau_1^{-1}(\tau_1(C^*)\cap M)\cap\tau_2^{-1}(\tau_2(C^*)\cap
    M). \SUBSET C^*\cap M\enspace.
  \end{equation}
  This chain is correct especially if
  $\tau_i(C^*)\SETMINUS\overline\alpha_k(0)\neq\emptyset$, $i=1,2$ and
  both set are $\alpha_k$-connected.

  In both cases the separation property follows.
\end{Beweis}

\begin{Lemma}\label{doppelpunkt1}
  It holds:
  \begin{enumerate}
  \item The set $\alpha_{n-1}(0)\SUBSET\Z^n$ contains no
    $\alpha_k$-double points for $0\le k\le n-1$.
  \item The set $\alpha_{k}(0)\SUBSET\Z^n$ contains no
    $\alpha_{n-1}$-double points for $0\le k\le n-2$.
  \end{enumerate}
\end{Lemma}

\begin{Beweis}
  1. Let $p$ be in $\alpha_{n-1}(0)$. Then, the point $p$ has the form
  $(0,\ldots,0,\pm1,0,\ldots,0)$. It cannot contain any $\pi$-neighbors
  $r=(r_1,\ldots,r_n)$ in $\alpha_{n-1}(0)$, because these satisfy
  \begin{equation}
    \sum_{i=1}^n |p_i-r_i|=1\enspace.
  \end{equation}
  The point $r$ cannot be 0 and satisfies:
  \begin{equation}
    \sum_{i=1}^n|r_i|=2\enspace.
  \end{equation}
  Therefore, no neighbor of $p$ can be contained in $\alpha_{n-1}(0)$ and 
  no $p$ exists, which satisfies the definition \ref{doppelpunkt}.

  2. We need to show, that for no $p\in\alpha_k(0)$ with $0\le k\le n-2$, exist two
  points $r\in\pi(0)\cap\alpha_{n-2}(p)$ and $q\in\pi(0)\cap\alpha_k(0)$ and a translation
  $\sigma$ with $\sigma(r)=0$ and $\sigma(q)=p$  such that $r\in\alpha_{n-1}(p)$.

  Assume for contradiction that such a configuration exists. Then, the two points 
  $q$ and $r$ are $\alpha_{n-1}$-adjacent. Therefore, it holds:
  \begin{equation}
    \sum_{i=1}^n|r_i-q_i|=1, |r_i-q_i|\le 1\mbox{ for } 1\le i\le
    n\enspace. 
  \end{equation}
  It follows the existence of a $j$ in $\{1,\ldots,n\}$ such that $r_j\neq
  q_j$ and $r_i=q_i$ for all other indices $i$. Furthermore, the point $q$ is in $\pi(0)$
  and it can be written as $(0,\ldots,\pm1,0,\ldots,0)$ with $q_l=\pm1$ and we know 
  that $q_l=r_l$. From $\sigma(r)=0$ it follows that $(-\sigma)(q)=p=(q_1-r_1,\ldots,q_n-r_n)$ 
  and therefore, the point $p$ has the form 
  \begin{equation}
    p=(0,\ldots,0,q_j-r_j,0,\ldots,0)\enspace.
  \end{equation}
  In addition, $r$ is an element of $\pi(p)$ and $|p_i-r_i|\le 1$ for all $1\le i\le n$.
  But this cannot be the case, since
  \begin{equation}
    |p_j-r_j|=|q_l-2r_l|=|-2r_l|=2\mbox{ since } r_l\neq 0\enspace.
  \end{equation}
  This contradicts the assumption and the Lemma is proven.
\end{Beweis}

\begin{Lemma}\label{doppelpunkt2}
  Given a pair $(\alpha_l\alpha_k)$ on $\Z^n$ with $n\ge 2$, the set 
  $\alpha_k(0)\SUBSET\Z^n$ contains $\alpha_l$-double points for all $0\le
  k\le n-2$ and $0\le l \le n-2$. 
\end{Lemma}

\begin{Beweis}
  Consider the point  $p=(1,1,0,\ldots,0)\in\alpha_k(0)$ with $k$ conforming
  the precondition.
  The point $q=(1,0,\ldots,0)$ is in $\pi(0)$ and the point 
  $r=(0,1,0,\ldots,0)$ is in $\pi(p)$.  
  In addition a translation $\sigma$ exists such that $\sigma(z)=r$ and $\sigma(p)=q$.

  Because of $q\in\alpha_l(r)$ for $0\le l\le n-2$, the Lemma is true.
\end{Beweis}

We now have all the tools in our hands to state the final Theorem on the good pairs
of cubical ajacencies. This Theorem gives us a complete characterization of this kind of
good pairs in $\Z^n$ for all dimensions $n$ at least 2.

\begin{Satz}
  A pair of cubical adjacencies $(\alpha_l,\alpha_k)$ in $\Z^n$ is a good pair, if
  \begin{enumerate}
  \item $k=n-1$ and $0\le l\le n-2$,
  \item $0\le k\le n-2$ and $l=n-1$. 
  \end{enumerate}
  There are no other good pairs of cubical adjacencies.
\end{Satz}

\begin{Beweis}
  We need to show that $\alpha_k(0)$ is a $(n-1)$-manifold in $\Z^n$ that contains
  no $\alpha_l$-double points. Lemma \ref{doppelpunkt1} gives the pairs 
  $(\alpha_l,\alpha_k)$ without double points.
  Corollary \ref{kub-surf} shows that $\alpha_k(0)$ is a $(n-1)$-manifold under $\alpha_l$,
  this is enough because of the invariance under translation of $\alpha_k$.
  And from Lemma \ref{doppelpunkt2} we know which pairs of cubical adjacencies have 
  double points.
\end{Beweis}

\subsection{The Khalimsky-Topology as Good Pair of Adjacencies}\label{section-khalimsky-top}

In this section we will show, that the notion of an Alexandrov-space 
and the graph-theoretic framework common to digital geometry may be put under a common
umbrella. We will see, that the Khalimsky-topology $\kappa_n$ on the set $\Z^n$ 
might be considered as a pair of adjacencies $(\kappa_n,\kappa_n)$, and that these
pairs a good ones.

Basing on Theorem \ref{alexandrov-partial-order} we may consider a graph structure
on $\Z^n$ given by the topology $\kappa_n$. We denote this graphical adjacency also with
$\kappa_n$. Also, remember the equations \ref{eq:closure1} and \ref{eq:closure2}.

\begin{Lemma}
  For any $p,q\in\Z^n$ holds: 
  \begin{equation}
    p\in\C_{\kappa_n}(q)\IFF\FORALL_{i=1}^n(p_i\ge q_i\mod{2}) :\IFF p\succeq q
  \end{equation}
  \begin{equation}
    p\in\U_{\kappa_n}(q)\IFF\FORALL_{i=1}^n(p_i\le q_i\mod{2}) :\IFF p\preceq q
  \end{equation}
\end{Lemma}

\begin{Beweis}
  This is Theorem 8 in Evako et al. \cite{evako}.
\end{Beweis}

The Khalimsky-adjacency $\kappa_n$ may now be represented in the following way:
\begin{equation}
  \kappa_n(p):=\left\{q\in\Z^n: \max_{i=1,\ldots,n}|q_i-p_i|=1,
  p\preceq q\lor q\preceq p \right\}\enspace.
\end{equation}

\begin{Lemma}
  For all $n\ge 1$ holds: $\pi\SUBSET\kappa_n$.  
\end{Lemma}

\begin{Beweis}
  Let $p,q$ be two points in $\Z^n$ such that $p\in\pi(q)$. By definition of $\pi$
  we have
  \begin{equation}
    \max_{i=1,\ldots,n}|p_i-q_i|=1\mbox{ and }\sum_{i=1}^n|p_i-q_i|\le 1
  \end{equation}
  Therefore, exactly one $i\in\{0,\ldots,n\}$ exists with $q_i=p_i+1$ or
  $q_i=p_i-1$. For all $j\in\{1,\ldots,n\}$, $j\neq i$ is $p_i=q_i$.
  We have:
  \begin{equation}
    \FORALL_{i=1}^n(p_i\le q_i\mod{2})\mbox{ or }
    \FORALL_{i=1}^n(p_i\ge q_i\mod{2})\enspace.
  \end{equation}
  And so, $p\in\kappa_n(q)$.
\end{Beweis}

We are not in the convenient position to find a reference point like 0 for the cubical
adjacencies. The next Lemma clarifies this fact.

\begin{Lemma}
  For each $p\in\Z^n$ exists a translation $\tau$ such that
  $\kappa_n(\tau(p))\neq \tau(\kappa_n(p))$.
\end{Lemma}

\begin{Beweis}
  By construction of the Khalimsky-topology this Lemma is obviously true:
  Let $\tau$ be any translation of the form $(0,\ldots,0,1,0,\ldots,0)$ in $\Z^n$.
  Then, $\tau(p)$ is either odd in a component where $p$ is even or vice versa.
  In both cases, the point $\tau(p)$ has a neighborhood different from the one
  of $p$.
\end{Beweis}

We are able to make some statements about the interaction of certain translations and
$\kappa_n$. 

\begin{Lemma}\label{kappa-translation}
  Let $p$ and $q$ be two points in $\Z^n$, $I=\{i: p_i=q_i\}$ and let
  $\tau$ be a translation with$|\tau(0)_i|\le 1$ for $i\in I$ and
  $\tau(0)_i=0$ otherwise. Then holds 
  \begin{equation}
    p\preceq q \IFF \tau(p)\preceq\tau(q)\enspace.
  \end{equation}
\end{Lemma}

\begin{Beweis}
  ($\IF$) Let $p\preceq q$. Then holds $p_j\le q_j\mod{2}$ for all
  $j\not\in I$. Because of $p_i=q_i$ we deduce $p_i\pm 1= q_i\pm 1\mod{2}$. Therefore,
  it holds $\tau(p)\preceq\tau(q)$.
  ($\ONLYIF$) Analog.
\end{Beweis}

\begin{Lemma}
  For all $p\in\Z^n$, $n\ge 2$, the set $\kappa_n(p)$ is
  $\kappa_n$-connected.
\end{Lemma}

\begin{Beweis}
The Lemma follows by Definition 4 and Theorem 11 in Evako
et al. \cite{evako}.
\end{Beweis}

From the proof of Theorem 11 in Evako et al. \cite{evako} we get

\begin{Lemma}\label{k:cube-connected}
  For all $p\in\Z^n$, $n\ge 2$, every $n$-cube, that contains points from 
  $\kappa_n(p)$, is $\kappa_n$-connected.\qed
\end{Lemma}

\begin{Lemma}\label{k:two-components}
  For all $p\in\Z^n$, $n\ge 2$, and all $q\in\kappa_n(p)$ the set
  $\omega(q)\SETMINUS\kappa_n(p)$ has exactly two $\kappa_n$-components
  $C_q$ and $D_q$.
\end{Lemma}

\begin{Beweis}
  Let $p$ and $q$ be the same as in the last Lemma.
  A $\kappa_n$-component of $\omega(q)\SETMINUS\kappa_n(p)$ is $\{p\}$, because $p$ 
  has in $\omega(q)$ only neighbors $\kappa_n(p)$. 
  We denote this component by $C_q$.

  Now define
  \begin{equation}
    \omega(q)_i := \left\{r\in\omega(q):|r_i-p_i|=2 \right\}
  \end{equation}
  We will show that this set is $\pi$-connected for all $1\le i\le n$
  We prove the result w.l.o.g. for $i=1$.
  
  The point 
  \begin{equation}
    r=(p_1+2,p_2,\ldots,p_n)
  \end{equation}
  is in $\omega(q)_1$. 
  Let $r'\neq r$ be any point in $\omega(q)_i$ and let $i\in\{2,\ldots,n\}$
  be the smallest index such that $r_i\neq r'_i$. 
  We construct a $\pi$-path from $r'$ to $r$.
  The point 
  \begin{equation}
    r''=(r_1,\ldots,r_i,r'_{i+1},\ldots,r'n)
  \end{equation}
  is a $\pi$-neighbor of $r'$, because, both points differ according to the choice 
  of $i$ only in the $i$-th coordinate by 1. If $r''=r$ the the path is constructed, 
  otherwise we iterate the algorithm with $r''$ in place of $r'$. After at most $n-1$ 
  steps the $\pi$-path is constructed.

  If the intersection of two sets $\omega(q)_i$ and $\omega(q)_j$ is non-empty, then 
  it is $\pi$-connected, too.

  Let $r$ be a point in the set
  \begin{equation}
    D_q:=\omega(q)\cap(\omega((p)\SETMINUS(\kappa_n(p)\cup\{p\}))\enspace.
  \end{equation}
  If $\omega(q)_i\neq\emptyset$ and $r_i=q_i$ for this 
  $i\in\{1,\ldots,n\}$, then $r_i$ is the $\pi$-neighbor of some point in $\omega(q)_i$.

  Otherwise, an $i$ exists such that $\omega(q)_i\neq\emptyset$. From
  $r_i\neq q_i$ follows $|r_i-q_i|=1$ and therefore $r_i=p_i$.
  We may define the point 
  \begin{equation}
    s=(r_1,\ldots,r_{i-1},q_i,r_{i+1},\ldots,r_n)\enspace.
  \end{equation}
  The points $r,s$ are in $\omega(p)$, so we have
  \begin{equation}
    \max_{i=1,\ldots,n}|r_i-p_i|=1\mbox{ and
    }\max_{i=1,\ldots,n}|s_i-p_i|=1\enspace.
  \end{equation}

  Since the point $r$ is no member of $\kappa_n(p)$, it follows:
  \begin{equation}
    \EXISTS_{j_1}(r_{j_1}>p_{j_1})\mbox{ and } \EXISTS_{j_2}(
    r_{j_2}<p_{j_2})\enspace.
  \end{equation}
  The indices $j_1$ and $j_2$ are distinct. From $r_i=p_i$
  follows, that $j_1,j_2$ are both dissimilar to $i$. 
  Therefore we have for $s$:
  \begin{equation}
    s_{j_1}=r_{j_1}>p_{j_1}\mbox{ and } s_{j_2}=r_{j_2}<p_{j_2}\enspace,
  \end{equation}
  which gives $s\not\in\kappa_n(p)$. Thereby,
  $s$ is in $D_q$ and $D_q$ is the second $\kappa_n$-component of the set
  $\omega(q)\SETMINUS\kappa_n(p)$.
\end{Beweis}

\begin{Lemma}\label{k:background-adjacency}
  For all $p\in\Z^n$ with $n\ge 2$ and any $q\in\kappa_n(p)$, all the points
  $r\in\kappa_n(p)\cap\kappa_n(q)$ are $\kappa_n$-adjacent to the sets $C_q$ and
  $D_q$ .
\end{Lemma}

\begin{Beweis}
  It obvious, that all points $r\in\kappa_n(p)\cap\kappa_n(q)$ are 
  $\kappa_n$-adjacent to the set $C_q=\{p\}$. So it remains to show, that $r$ is also
  $\kappa_n$-adjacent to $D_q$.

  Case 1: For some index $i$ in $\{0,\ldots,n\}$ holds that $r_i=q_i$ and the set 
  $\omega(q)_i$ is not empty. Then, the point $r$ is $\pi$-adjacent to $D_q$.

  Case 2: It is  $r_i\neq q_i$ for all $i$ such that $\omega(q)_i\neq\emptyset$. 
  Consider the set $I=\{i:\omega(q)_i\neq\emptyset\}$. We show that the point
  \begin{equation}
    s\mbox{ with } 
    s_i=\left\{
    \begin{array}{rcl}
      q_i&\mbox{if}& i\in I \\
      r_i&\mbox{if}& i\in\{1,\ldots,n\}\SETMINUS I
    \end{array}
    \right.
  \end{equation}
  is no member of $\kappa_n(p)$ under this preconditions. Since $s$ can be identified as
  $\kappa_n$-adjacent to $r$, the point $r$ is $\kappa_n$-adjacent to $D_p$.

  The point $s$ is distinct from $q$ by definition of $I$ and  $r\neq p$. We know that $r_i=p_i$ 
  for $i\in I$ because of $r_i\neq q_i$ and $|r_i-p_i|<2$ and $r$ is a member of $\omega(q)$. 
  The set $\omega(q)_i$ is non-empty if and only if $q_i\neq p_i$. Therefore, a translation
  $\tau$ exists such that
  \begin{equation}
    \tau(0)_i=\left\{
    \begin{array}{rcl}
      \pm1&\mbox{if}&i\in\{1,\ldots,n\}\SETMINUS I\\
      0&\mbox{if}&i\in I
    \end{array}\right.
  \end{equation}
  We may choose $\tau$ such that $\tau(q)=p$ and $\tau(s)=r$.

  The point $q$ is in $\kappa_n(p)$. Suppose w.l.o.g. that $q\preceq
  p$. Therefore, we get $q_i< p_i=r_i\mod{2}$ for $i\in I$. 
  Then follows $q\preceq r$ with $r\in\kappa_n(q)$. By definition of
  $s$ and the fact $r\neq p$, it holds that $q\preceq s$ and by Lemma
  \ref{kappa-translation} we get 
  \begin{equation}
    \tau(q)=p\preceq r=\tau(s)\enspace.
  \end{equation}
  So we can find a $j$ with $s_j=r_j>p_j\mod{2}$ and $j\not\in I$. But at the same time
  $q_i=s_i<p_i\mod{2}$ for all  $i\in I$. Therefore the point $s$ cannot be contained in
  $\kappa_n(p)$. We have to show that $r$ and $s$ are $\kappa_n$-neighbors: For $i\not\in I$ 
  we have $r_i=s_i$ and for $i\in I$ it holds
  \begin{equation}
    s_i = q_i < p_i = r_i \mod 2
  \end{equation}
  and so follows $s\preceq r$ which means $s\in\kappa_n(r)$.
\end{Beweis}

For the proof of the separation property we consider the set
$\overline\kappa_n(p)=\kappa_n(p)\cup\{p\}$ for all $p\in\Z^n$, $n\ge 2$
instead of $\kappa_n(p)$. This is reasonable, since the point $p$ lies in no
separable component of the complement of $\kappa_n(p)$ in $\Z^n$. If we have the
result for the modified set we may easily translate it for the original one.

\begin{Lemma}
  Let $C$ be any $k$-cube $\omega(p)\cup\{p\}$, $0\le k\le n$ and let $q$ be the point in 
  $C$ with minimal $\pi$-distance\footnote{The $\pi$-distance of two points $p$ and $q$ is 
    the infimum over the length of all $\pi$-paths from $p$ to $q$.} to $p$. 
  For $q\preceq p$ and all $q'\in C\SETMINUS\{q\}$ holds 
  \begin{equation}
    q'\preceq q\IFF q'\preceq p\enspace.
  \end{equation}
  
  An analog claim holds for $q\succeq p$.
\end{Lemma}

\begin{Beweis}
  ($\IF$) This direction of the proof follows by transitivity of the partial order $\preceq$. 
  ($\ONLYIF$) The points $q'$ and $q$ are contained in the same $k$-cube $C$ and it holds that
  \begin{equation}
    \sum_{i=1}^n|q_i-p_i|=k<l=\sum_{i=1}^n|q'_i-p_i|\enspace.
  \end{equation}
  After rearranging the coordinates of $q,q'$ and $p$, we get
  \begin{equation}
    q'=(q_1,\ldots,q_{k},q'_{k+1},\ldots,q'_l,p_{l+1},\ldots,p_n)
  \end{equation}
  \begin{equation}
    q=(q_1,\ldots,q_{k},p_{k+1},\ldots,p_l,p_{l+1},\ldots,p_n)\enspace.
  \end{equation}
  From $q'\preceq p$ now follows $q\preceq p$ by the definition of
  $\preceq$. 
\end{Beweis}

\begin{Lemma}
  Let $C$ be a $k$-cube $2\le k\le n$ and $q$ be a point in $C$ with minimal $\pi$-distance to 
  $p$ and $q\preceq p$. For all $q'\in\pi(q)\cap C$ holds $q'\preceq q$ if and only if $C$ is
  contained in the set $\overline\kappa_n(p)$.

  An analog claim holds for $q\succeq p$.
\end{Lemma}

\begin{Beweis}
  ($\ONLYIF$) Since $C\SUBSET\overline\kappa_n(p)$, all the points $q'\in C\cap\pi(q)$ are
  in $\kappa_n(p)$. Therefore, they satisfy $q'\preceq q$ or $q'\succeq q$. 
  If there exists a $q'\preceq q$ and a $q''\succeq q$, so we have
  \begin{equation}
    \EXISTS_{i_1}(q'_{i_1}<q_{i_1}\mod{2})\mbox{ and
    }\EXISTS_{i_2}(q''_{i_2}>q_{i_2}\mod{2})\enspace.
  \end{equation}
  The point $q'''=\tau_{i_1}\tau_{i_2}(q)$ then satisfies
  \begin{equation}
    \EXISTS_{i_1}(q'''_{i_1}<q_{i_1}\mod{2})\mbox{ and
    }\EXISTS_{i_2}(q'''_{i_2}>q_{i_2}\mod{2})\enspace.
  \end{equation}
  Therefore holds $q'''\not\preceq q$ and $q'''\not\succeq q$ and the point $q'''$
  no member of $\overline\kappa_n(p)$. So for all points $q'\in C\cap\pi(q)$ 
  the relation $q'\preceq q$ holds.

  ($\IF$) We prove by induction on $k$. 
  In the case $k=2$ holds $q\preceq p$ and for all $q'\in C\cap\pi(q)$ holds 
  $q'\preceq q\preceq p$. The two $\pi$-neighbors $q_1$ and $q_2$ of $q$ in $C$ are in 
  $\overline\kappa_n(p)$. This means that
  $q_1=\tau_1(q)\preceq q$ and $q_2=\tau_2(q)\preceq q$. 
  Therefore, we have
  \begin{equation}
    \tau_1(\tau_2(q))\preceq \tau_1(q)\preceq p\enspace.
  \end{equation}
  We conclude, that $C$ is contained in $\overline\kappa_n(p)$.

  For the induction step $k>2$ we let $C=C'\cup\tau(C')$ for certain $(k-1)$-cubes $C',\tau(C')$
  and a translation $\tau$. Let $q$ be in $C$ w.l.o.g.
  Since all the points $q'$ in $\pi(q)\cap C$ satisfy the relation
  $q'\preceq q$, the $(k-1)$-cube $C'$ has to be contained by induction hypothesis in 
  $\overline\kappa_n(p)$. For all $q''\in C'$ holds $q''\preceq q$. 
  Therefore, by Lemma \ref{kappa-translation}, we find for all $\tau(q'')\in\tau(C')$:
  \begin{equation}
    \tau(q'')\preceq\tau(q)\preceq q\preceq p\enspace.
  \end{equation}
  It follows that $C\SUBSET\overline\kappa_n(p)$.
\end{Beweis}

\begin{Folg}
  Let $C$ be a $k$-cube, $2\le k\le n$ and $q$ be the point with
  minimal $\pi$-distance to $p$. Then, all the subcubes $C'$ of $C$ such that
  $q'\preceq q\preceq p$ or $q'\succeq q\succeq p$ for all $q'\in
  C'\cap\pi(q)$, are contained in $\overline\kappa_n(p)$.

  For $q\neq p$ only one of these cases applies.\qed
\end{Folg}

\begin{Lemma}\label{kappa-trenn1}
  The set $\overline\kappa_n(p)$ has the separation property under the pair 
  $(\kappa_n,\kappa_n)$ for any cube $C\SUBSET(\omega(p)\cup\{p\})$ with $p\not\in C$.
\end{Lemma}

\begin{Beweis}
  We consider three cases. The first case is, that $C$ is contained in
  $\overline\kappa_n(p)$. The separation property is obviously satisfied in this case.

  Case 2: Let the $k$-cube $C$ be of the form $C'\cup\tau(C')$ with $C'$ a $(k-1)$-cube
  contained in $\overline\kappa_n(p)$. In this case the set $\tau(C')$
  contains no points $q'$ in $\overline\kappa_n(p)$, since otherwise these points would
  satisfy $q'\le q\le p$. Particularly, the point $\tau(q)$ is not in $\overline\kappa_n(p)$. 

  Every $(k-2)$-cube $C''$ with $C''\cap\overline\kappa_n(p)$ is in $C'$.
  Then, the set $\tau_1(C'')\SUBSET C'$ is also contained in
  $\overline\kappa_n(p)$. Therefore the separation property holds in $C$.

  Case 3: There is only one $(k-2)$-subcube $C'$ of $C$ that contains all points of 
  $C\cap\overline\kappa_n(p)$. Then we get
  $\tau_1\tau_2(C')\cap\overline\kappa_n(p)=\emptyset$. Therefore it holds
  \begin{equation}
    (\tau_1\tau_2)^{-1}(\tau_1\tau_2(C')\cap\overline\kappa_n(p))
    \SUBSET (\tau_1^{-1}(\tau_1(C')\cap\overline\kappa_n(p))) \cap
    (\tau_2^{-1}(\tau_2(C')\cap\overline\kappa_n(p)))\enspace.
  \end{equation}
  And so, the separation property holds.
\end{Beweis}

\begin{Lemma}\label{kappa-trenn2}
  The set $\overline\kappa_n(p)$ has the separation property for the pairs
  $(\kappa_n,\kappa_n)$ for cubes $C\SUBSET\omega(p)\cup\{p\})$ with $p\in C$.
\end{Lemma}

\begin{Beweis}
  Case 1: The separation property is satisfied for $C\SUBSET\overline\kappa_n(p)$.

  Case 2: For a $k$-cube $C$ of the form $C'\cup\tau(C')$ such that 
  $C'\SUBSET\overline\kappa_n(p)$ only the point $\tau(p)$ is in
  $\overline\kappa_n(p)$, because, if for all $q\in C'$ the relation
  $q\succeq p$ is true, then it holds for $\tau(q)\in\tau(C')$ that
  \begin{equation}
    \tau(q)\succeq\tau(p)\preceq p\enspace.
  \end{equation}
  Since $\tau(p)$ has minimal $\pi$-distance to $p$ in $\tau(C')$, none of the aforementioned 
  $\tau(q)$ can be contained in $\kappa_n(p)$.

  Let $C''\SUBSET C'$ be any $(k-2)$-cube. Then, the set $C''\cap\overline\kappa_n(p)$ is
  maximal with respect to inclusion in $C$. In turn, the set
  $\tau_1(C'')\SETMINUS \overline\kappa_n(p)$ is empty and the separation property holds 
  for $C$.

  Case 3: Consider the $k$-cube $C=C'\cup\tau_1(C')\cup\tau_2(C')\cup\tau_1\tau_2(C')$ 
  and let $C'\cap\overline\kappa_n(p)$ be maximal with respect to inclusion. 
  Since we are not in case 2, we have
  $\tau_1(C')\SETMINUS\overline\kappa_n(p)\neq\emptyset$. The $(k-1)$-cube $C'$
  has a $l$-subcube, $0\le l<k-1$, that is contained in $\overline\kappa_n(p)$,
  the point $p$ has to be in $C'$ liegen. Now, either all points $q\in C'$ are in
  relation $q\preceq p$ or they satisfy $q\succeq p$. 
  W.l.o.g. we use the first relation.

  All the points in $q\in\tau_i(C')$, $i=1,2$, are in the relation $q\succeq p$,
  since otherwise, we had $C'\cup\tau_i(C')\SUBSET\overline\kappa_n(p)$.
  Now, we have $\tau_1\tau_2(p)\succeq\tau_1(p)\tau_2(p)\succeq p$. Likewise, all the 
  translations $\tau$, that generate the $(k-1)$-cube $C'$, satisfy by Lemma
  \ref{kappa-translation}:
  \begin{equation}
    \tau_1\tau_2(\tau(p))\succeq \tau_1(\tau(p)),\tau_2(\tau(p))\succeq \tau(p)
  \end{equation}

  Therefore, only the points $\tau_1\tau_2(\tau(p))$ and $\tau_i(\tau(p))$, $i=1,2$ 
  are in $\overline\kappa_n(p)$, if $\tau(p)\succeq p$ holds. So we have 
  \begin{equation}
    (\tau_1\tau_2)^{-1}(\tau_1\tau_2(C')\cap\overline\kappa_n(p))
    \SUBSET (\tau_1^{-1}(\tau_1(C')\cap\overline\kappa_n(p))) \cap
    (\tau_2^{-1}(\tau_2(C')\cap\overline\kappa_n(p))) 
  \end{equation}
  and the separation property holds in $C$.  
\end{Beweis}

\begin{Folg}\label{k:separation-property}
  The set $\overline\kappa_n(p)$ has the separation property under the pair $(\kappa_n,\kappa_n)$.
\end{Folg}

\begin{Beweis}
  The claim follows from the Lemmata \ref{kappa-trenn1} and
  \ref{kappa-trenn2} for cubes $C\SUBSET\omega(p)\cup\{p\}$. 

  For any cube $C$ that is not contained in $\omega(p)\cup\{p\}$, the separation property 
  holds, because $C$ has the form $C'\cup\tau_1(C')\cup\tau_2(C')\cup\tau_1\tau_2(C')$ 
  and the set $\tau_1\tau_2(C')\cap\overline\kappa_n(p)$ is always empty, since $C'\SUBSET
  \omega(p)\cup\{p\}$ is true if we maximize the set $C'\cap\overline\kappa_n(p)$ with 
  respect to inclusion. In the case
  $\tau_1(C)\SETMINUS\overline\kappa_n(p)=\emptyset$ the separation property holds trivially.
  For $\tau_1(C)\SETMINUS\overline\kappa_n(p)\neq\emptyset$ this is also true because of
  \begin{equation}
    (\tau_1\tau_2)^{-1}(\tau_1\tau_2(C')\cap\overline\kappa_n(p))=\emptyset
    \SUBSET (\tau_1^{-1}(\tau_1(C')\cap\overline\kappa_n(p))) \cap
    (\tau_2^{-1}(\tau_2(C')\cap\overline\kappa_n(p)))\enspace.
  \end{equation}
  And so the separation property holds again.
\end{Beweis}

\begin{Satz}\label{kappa-mannigfaltigkeit}
  For all $p\in\Z^n$, $n\ge 2$ the set $\kappa_n(p)$ is a
  $(n-1)$-manifold. 
\end{Satz}

\begin{Beweis} 
The first three properties of a digital $(n-1)$-manifold
are shown in the Lemmata \ref{k:cube-connected} to \ref{k:background-adjacency} and the
separation property is proven in Corollary \ref{k:separation-property}. 
\end{Beweis}

\begin{Lemma}\label{kappa-doppel}
  Given the pair $(\kappa_n,\kappa_n)$ on $\Z^n$, $n\ge 2$, and any point $p\in\Z^n$, the set 
  $\kappa_n(p)$ contains no $\kappa_n$-double points.
\end{Lemma}

\begin{Beweis}
  Assume for contradiction, we have the points $z\in\kappa_n(p)$, $q\in\kappa_n(p)\cap\pi(z)$ and
  $r\in\kappa_n(z)\cap\pi(p)$, and $q=\sigma(p)$ and $z=\sigma(r)$ for a simple
  translation $\sigma$.
  
  The point $z$ is in $\kappa_n(p)$ and so we have $z\preceq p$ or $p\preceq
  z$. We consider w.l.o.g. the case $z\preceq p$. 
  We have exactly one $i\in\{1,\ldots,n\}$ such that
  \begin{equation}
    z_i=r_i<p_i\mod{2}\enspace.
  \end{equation}
  Therefore, it holds that
  \begin{equation}
    q_i=p_i>r_i\mod{2}\enspace.
  \end{equation}
  Furthermore, we can find a $j\in\{1,\ldots,n\}$ such that 
  \begin{equation}
    z_j=q_j<p_j=r_j\mod{2}\enspace.
  \end{equation}
  It follows that $q_i>r_i\mod{2}$ and $q_j<r_j\mod{2}$. Therefore neither 
   $q\preceq r$ nor $r\preceq q$ may be true. This contradicts the assumption that 
  $q\in\kappa(p)$ and so no double points may occur.
\end{Beweis}

\begin{Satz}
  The pair $(\kappa_n,\kappa_n)$ is a good pair on $\Z^n$ for
  all $n\ge 2$.
\end{Satz}

\begin{Beweis}
  The proof follows with Theorem \ref{kappa-mannigfaltigkeit} and
  Lemma \ref{kappa-doppel}.
\end{Beweis}

\section{Conclusions}

We have shown that the cubical adjacencies and the khalimsky-topology give good pairs.
This was already known, for instance G.T.~Herman proved this in his book \cite{herman}.
The difference here is, that our theory resembles more closely the euclidean case and surfaces
are really subsets of the given space. We also could give a slight unification of the 
topological with the graph-theoretic setting, although this was already present in the
disguise of Alexandrov-spaces, for these have an graph-theoretic interpretation via
partial orders. It is possible to give proofs for other adjacency relations to be good 
pairs, for instance the hexagonal adjacencies also give good pairs, as G.T. Herman 
shows in the same book. It may be also possible to give good pairs of more complicated 
adjacency relations, but then, the proofs might tend to get even more technical than 
the ones we saw we saw in this paper.

%%%%%%%%%%%%%%%%%%%%%%%%%%%%%%%%%%%%%%%%%%%%%%%%%%%%%%%%%%%%%%%%%%%%%%
%%
%% Literaturliste
%%

\end{document}